\documentclass[conference]{IEEEtran}
\IEEEoverridecommandlockouts
\usepackage{cite}
\usepackage{amsmath,amssymb,amsfonts}
\usepackage{algorithmic}
\usepackage{graphicx}
\usepackage{graphics}
\usepackage{textcomp}
\usepackage[export]{adjustbox}
\usepackage{multicol}
\usepackage{tabularx}
\usepackage{color}

\def\BibTeX{{\rm B\kern-.05em{\sc i\kern-.025em b}\kern-.08em
    T\kern-.1667em\lower.7ex\hbox{E}\kern-.125emX}}
\begin{document}

\title{Training Deep Neural Networks with Different Datasets In-the-wild: The Emotion Recognition Paradigm\\
}

\author{\IEEEauthorblockN{ Dimitrios Kollias}
\IEEEauthorblockA{\textit{Department of Computing} \\
\textit{Imperial College London}\\
United Kingdom \\
dimitrios.kollias15@imperial.ac.uk}
\and
\IEEEauthorblockN{Stefanos Zafeiriou}
\IEEEauthorblockA{\textit{Department of Computing} \\
\textit{Imperial College London}\\
United Kingdom \\}
\&\\
\IEEEauthorblockA{\textit{Center for Machine Vision and Signal Analysis} \\
\textit{University of Oulu}\\
Finland \\
s.zafeiriou@imperial.ac.uk}
}

\maketitle

\begin{abstract}
A novel procedure is presented in this paper, for training a deep convolutional and recurrent neural network, taking into account both the available training data set and some information extracted from similar networks trained with other relevant data sets. This information is included in an extended loss function used for the network training, so that the network  can have an improved performance when applied to the other data sets, without forgetting the learned knowledge from the original data set. Facial expression and emotion recognition in-the-wild is the test bed application that is used to demonstrate the improved performance achieved using the proposed approach. In this framework, we provide an experimental study on categorical emotion recognition using datasets from a very recent related emotion recognition challenge.  
\end{abstract}

\begin{IEEEkeywords}
deep neural network training; classification; clustering internal representations; extended loss function; domain adaptation; transfer learning; emotion recognition in-the-wild;  . 

\end{IEEEkeywords}

\section{Introduction}
Many real life problems are represented by a variety of data sets which may possess different characteristics. In such cases learning to classify correctly one data set does not generalize well in the other sets. Emotion recognition based on facial expressions is such a problem, due to the variations in expression of emotions among different persons, as well as to the different ways of labeling emotional states by different annotators. It would be of great interest if we could use a training data set to design a deep neural network (DNN), taking into account some knowledge about another set, so as to improve generalization of the network when applied to the other data set, without forgetting the original knowledge of it.

In this paper we deal with categorical emotion recognition based on facial expressions, treated as a classification problem in the seven primary emotional states, i.e., happiness, anger, fear, disgust, sadness, surprise and neutral state. Our approach can be extended to the dimensional emotion recognition problem as well, through discretization of the 2D continuous Valence Arousal Space \cite{plutchik1980emotion}.
Moreover, we focus on emotion recognition in-the-wild, i.e., on recognizing emotions expressed in real life, uncontrolled environments \cite{zafeiriou},\cite{zafeiriou1},\cite{dhall2017individual}.

A related Grand Challenge, the EmotiW one \cite{dhall2017individual}, has been constantly organized during the last five years, providing the AFEW data sets, consisting of videos showing persons expressing their emotions in real life. The Challenge provide training and validation data sets for designing emotion recognition approaches and test data for evaluating the performance of these approaches. 

Since the data are in the form of video sequences, we focus on Convolutional and Recurrent Neural Network (CNN-RNN) architectures. CNN-RNNs have achieved the best performances in all recent contests \cite{dhall2015video},\cite{dhall2017individual}, \cite{dhall2016emotiw}.

\begin{figure*}[t]
\centering
\adjincludegraphics[height=3cm,trim={.50\width} {.44\totalheight} {.57\width} {.44\totalheight}]{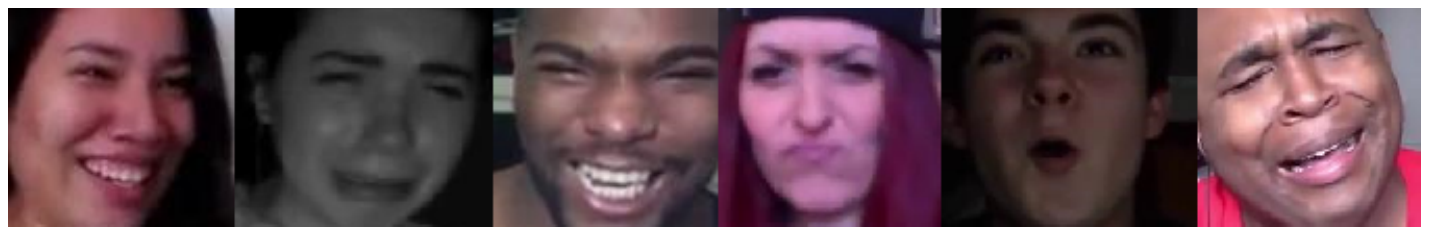} 
\caption{Video frames extracted from the Aff-Wild Dataset}
\label{rest}
\end{figure*}

Deep Convolutional Neural Networks (CNNs) \cite{krizhevsky2012imagenet},\cite{lecun2010convolutional}, \cite{bengio2009learning} include convolutional layers with feature maps composed of neurons with local receptive fields and shared weights and pooling layers, generating condensed representations. The resulting strength is the automatic hierarchical generation of rich internal representations, which are fed next to fully connected layers, for classification, or prediction, purposes. 

Recurrent neural networks (RNNs) have the ability to model time varying asynchronous patterns in audio and video \cite{yao2015capturing}. Since emotion events do not appear in a single frame, RNNs can follow and capture the events' sequential evolution.  
They include hidden layer(s) with long time dependencies. Their hidden units are functions of inputs and of hidden states, with their input being image pixel values, or features extracted from the images. Using the neuron Long Short-Term Memory (LSTM) model, one can overcome the backpropagation vanishing effect by introducing appropriate gates (input, update, output) and a cell state. BLSTMs are bidirectional LSTMs \cite{byeon2015scene}, processing input data in two directions. Other variants of LSTMs are the Gated Recurrent Units (GRUs) \cite{chung2014empirical} that have two gates (forget, update) rather than three.

It should be mentioned that there is a big discrepancy between the EmotiW training and validation data sets. Different annotation strategies have been used in them, based on weak annotation labeling and clustering. As a consequence, there are different annotation behaviors and biases between the two sets. it should be mentioned that the generalization accuracy when training a network with either of these sets and testing on the other set is around 40\%.  The test data characteristics are close to the validation data.

In the paper we consider the problem of using the Emotiw 2017 training data and some knowledge about the validation data, to design a deep neural architecture with improved generalization accuracy on the validation data set. We propose a domain adaptation approach, by generating respective representations from the two datasets, i.e., the training and validation ones in our case, and by then trying to match the statistical distribution of these representations.  As a consequence, a classification scheme, trained on the first data set, which would take into account the statistics of representations of the second dataset, would be able to improve its performance when applied to the latter one. Evidently, our first data set is the EmotiW 2017 training data and the second data set includes the respective validation data. 

We realize these representations, by extracting and using internal features generated by deep convolutional and recurrent neural (CNN-RNN) architectures, separately trained with the validation and with the training EmotiW data sets.  In this framework we propose and use a new loss function for training the CNN-RNN system with the EmotiW training set, including minimization of the difference in the statistics produced by this system and a respective one applied to the validation set. 

The paper is organized as follows. Section II describes the facial expression and emotion recognition in-the-wild problem, focusing on categorical emotion recognition, and particularly, on classifying audiovisual data in one of the seven basic emotion categories. Section III presents the proposed approach, describing the new error function adopted in training, as well as a new formulation extracting internal features and representations from trained CNN-RNN architecture. Section IV provides the experimental study illustrating the performance obtained through the proposed novel approach for categorical emotion recognition. Conclusions and reference to planned future developments are provided in Section V of the paper.

\section{Emotion Recognition in-the-wild}

A common way to represent emotions is through the categorical model, which uses the six basic emotion categories defined by \cite{ekman2013emotion}, i.e., happiness, sadness, anger, disgust, surprise and fear, as well as the neutral category to represent human behaviors. This is due to several psychophysical experiments suggesting that the perception of emotions by humans is categorical \cite{ekman1997face}.

Another way to think about human emotions is through the dimensional model \cite{russell1978evidence} \cite{whissel1989dictionary}. This model shows that emotions can be distributed on a two dimensional circular space which contains valence and arousal dimensions, with the center representing neutral valence and a medium level of arousal. Emotional states are represented in this model as any level of valence and arousal, or at a neutral level for one or both of them. 

Both models have been recently used by various approaches targeting what is called emotion recognition in-the-wild, meaning recognizing emotions expressed by different persons in their everyday life, i.e., in uncontrolled environments. New datasets have been generated, by aggregating and annotating, according to one of these models, videos from YouTube and Movies \cite{zafeiriou},\cite{dhall2017individual} and used in Emotion Recognition Challenges. Fig. \ref{rest} shows some examples of image frames, taken from the Aff-Wild Database \cite{zafeiriou}, showing facial expressions captured in-the-wild.  

In the following we focus on the EmotiW 2017 Challenge, which is the most recent one targeting categorical emotion recognition in-the-wild. This challenge was the fifth in a series, starting on 2013 \cite{dhall2013emotion}, with all of them focusing on the topic of emotion recognition from audio-visual data in (real world) uncontrolled conditions. 

\begin{figure}[!ht]
\begin{tabularx}{\linewidth}{>{\hsize=1.2\hsize}X
                             >{\hsize=1.2\hsize}X}
        \includegraphics[width=\hsize,valign=t]{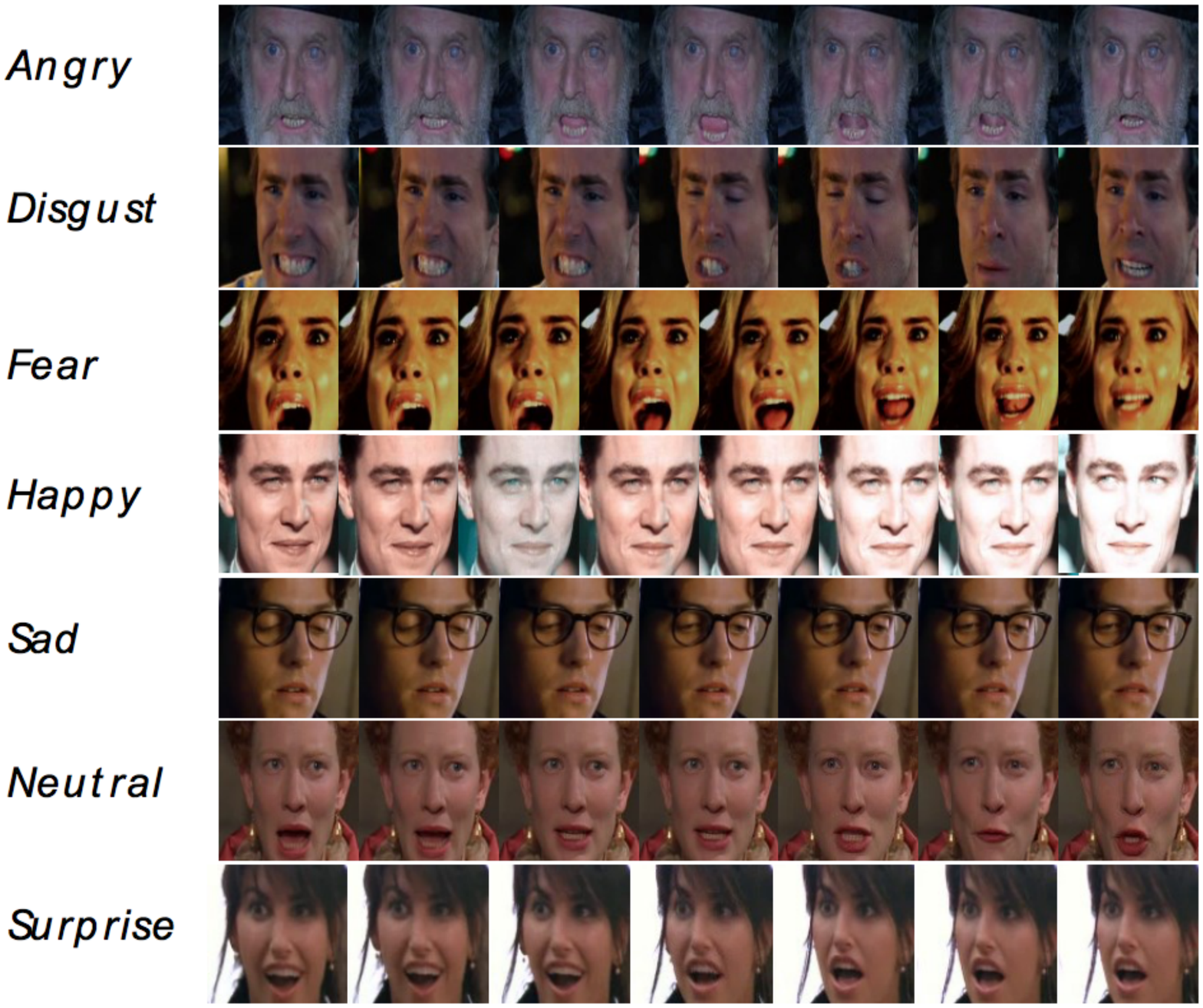}
\caption{Parts of Image Sequences from the AFEW dataset}
\label{emotiw_ims}
\end{tabularx}
\end{figure}

The series of EmotiW challenges make use of data from the Acted Facial Expression in-the-wild (AFEW) dataset. This dataset is a dynamic temporal facial expressions data corpus consisting of close to real world scenes extracted from movies and reality television shows. In total it contains 1809 videos. The whole dataset is split into three sets: training set (773 video clips), validation set (383 video clips) and test set (653 video clips). It should be emphasized that both training and validation sets are mainly composed of real movie records, however 114 out of 653 video clips in the test set are real TV clips, increasing, therefore, the difficulty of the challenge. The number of subjects is more than 330, aged 1-77 years. The annotation is according to 7 facial expressions (Anger, Disgust, Fear, Happiness, Neutral, Sadness and Surprise), as shown in Fig. \ref{emotiw_ims}. The challenges focus on audio-video classification of each clip into the seven basic emotion categories.

\section{The Proposed Approach}\label{sec3}

Let us assume that we have two datasets that include videos of different persons expressing their emotions in-the-wild; these have been annotated with reference to the seven basic emotion categories model. We will use the one dataset for training a Convolutional and Recurrent Neural Network (CNN-RNN) architecture (let us call it NetA) and the other for testing NetA's performance on it. This is different from transfer learning, which encompasses catastrophic forgetting, i.e., deterioration of the performance of NetA on its original training set,  after being fine-tuned with the new data set.

The proposed approach starts with the latter, test dataset. We train another CNN-RNN architecture (let us call it NetB) with these data, achieving the best possible emotion recognition accuracy. Such a network consists of the convolutional part, including one or more fully connected layers, using a ReLU type of neuron activation units, followed by the recurrent network part, which contains one or more layers, with B-LSTM, or GRU types of neurons. 

Let us now focus on the outputs of the last layer of the trained RNN part of NetB. In fact, it is through these output values that the network produces its final outputs, corresponding to the seven emotion categories. Assuming that a satisfactory performance is obtained by training NetB, it would be desirable to have NetA – when trained with its respective data set - generating output values, in its last before the output layer, which are close to the ones produced by NetB. If this happened, this would mean that training of NetA also managed to bring its own outputs closer to the ones generated by NetB. This is a desirable task, since both NetA and NetB target recognition of persons'\ emotions in different, randomly selected environments.

Our proposed approach is based on clustering the above-mentioned extracted internal representations of NetB into  seven clusters, corresponding to the targeted emotion categories and using the derived cluster centers as desired outputs for the respective representations generated at the corresponding layer during the training of NetA.     

In particular, let us denote, for the \textit{m}-th input sample, by 
\begin{equation} \label{eq:1}
U_m = \{ u_m^1,... ,u_m^n \}
\end{equation}
a vector with the CNN-RNN last (before the output) hidden layer neurons’ outputs, assuming it contains \textit{n} neurons. We perform clustering of the U values in seven clusters, corresponding to the seven basic emotion categories. The \textit{k}-means algorithm \cite{lloyd1982least} can be used for this task. 

Let 
\begin{equation} \label{eq:2}
Z_i = \{ z_i^1,...,z_i^n \}
\end{equation}
denote the seven cluster centroids (\textit{i} = 1, ..., 7), in the \textit{n}-dimensional representation space. These centroids, with their labels, constitute the information used in the proposed approach for adapting the training of NetA towards a clustered representation related to the best performance of NetB on its respective data set.

In the following, we introduce the above centroids, in the form of additional desired responses during training of NetA.
Let us denote by
\begin{equation} \label{eq:3}
O_m = \{ o_m^1,...,o_m^7 \}
\end{equation}
a vector with the seven NetA outputs, also corresponding to the basic emotion categories and  by
\begin{equation} \label{eq:4}
X_m = \{ x_m^1,...,x_m^n \}
\end{equation}
a respective vector with the NetA last (before the output) hidden layer neurons’ outputs. In (\ref{eq:2})-(\ref{eq:4}), \textit{m} also denotes the \textit{m}-th training input data sample of NetA.

Originally, training of NetA is done through minimization, with respect to the network weights, of the mean squared error between NetA outputs $O_m$ and desired outputs, say $D_m$,  defined as follows:
\begin{equation} \label{eq:5}
E_a = \frac{1}{7M} \sum_{m=1}^{M} \sum_{i=1}^{7} (d_m^i-o_m^i)^2 
\end{equation}
for \textit{m} covering all M data samples and the squared difference in (\ref{eq:5}) 
being computed between the respective seven components of $D_m$ and $O_m$.

In our formulation, we include a second term in error minimization, derived as follows. Let us define, using (\ref{eq:2}) and (\ref{eq:4}), for \textit{i}= 1,..,7, vector $Y_m^i$ and the values $V_m^i$:
\begin{equation} \label{eq:6}
Y_m^i = Z_i - X_m = \{ z_i^1-x_m^1,...,z_i^n-x_m^n \}
\end{equation}

\begin{equation} \label{eq:7}
V_m^i = Y_m^i* (Y_m^i)^T
\end{equation}
where \textit{T} denotes the transpose of the vector.

We request minimization of the $V_m^i$ value corresponding to the desired output category and maximization of the rest $V_m^i$ values corresponding to the other categories.

To achieve this, we normalize these $V_m^i$ values, either using a softmax \textit{f} function, or linearly, by dividing each one with the sum of all of them. Moreover, we subtract each normalized value from unity, so that the minimum distances correspond to maximum output values. 

This leads us to define a new Mean Squared Error criterion, between the computed values and the corresponding desired outputs $D_m$, as follows:
\begin{equation} \label{eq:8}
E_b = \frac{1}{7M} \sum_{m=1}^{M} \sum_{i=1}^{7} (d_m^i-[1-f(V_m^i)] )^2 
\end{equation}
where, similarly to (\ref{eq:5}) the squared difference is computed over all training data samples.

We target minimization of both Error Criteria in (\ref{eq:5}) and (\ref{eq:8}) during training of NetA, through the following combined Loss Function,
\begin{equation} \label{eq:9}
E_{tot} = \lambda E_a + (1-\lambda) E_b                               
\end{equation}
by selecting appropriate values for the tuning parameter \textit{$\lambda$} in [$0 $, $1 $]. 

\begin{figure}[!ht]
\begin{tabularx}{\linewidth}{>{\hsize=1.2\hsize}X
                             >{\hsize=1.2\hsize}X}
        \includegraphics[width=\hsize,valign=t]{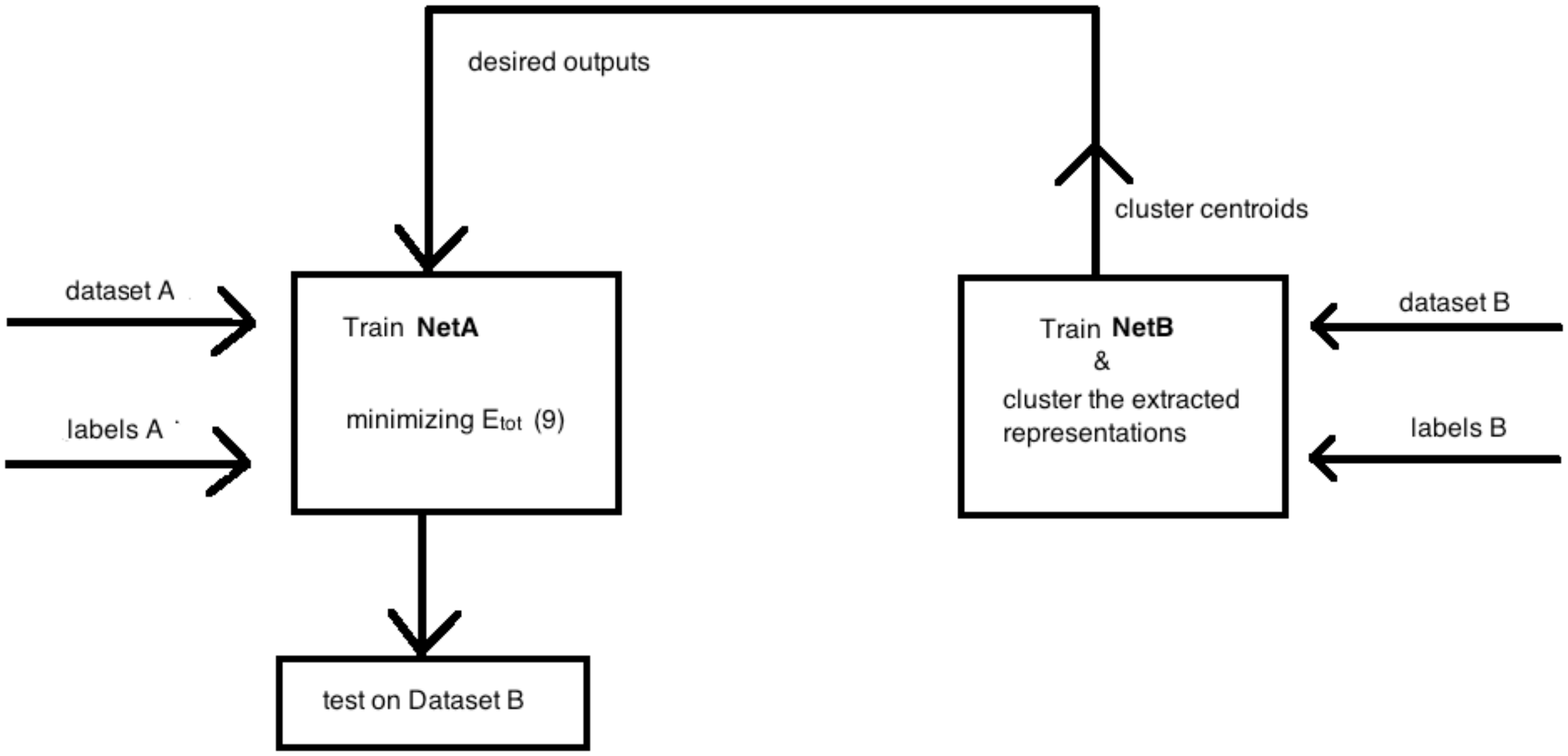}
\caption{The proposed procedure for generating different facets of a Deep Neural Architecture designed for Emotion Recognition in-the-wild}
\label{fig3}
\end{tabularx}
\end{figure}

Fig. \ref{fig3} shows the proposed approach, which can be repeated with any new NetB, providing different facets of NetA that can be used for analyzing corresponding datasets in-the-wild.

In the experimental study which follows, we will illustrate that, using (\ref{eq:9}) to train a CNN-RNN network for categorical emotion recognition, provides improved performance of the trained NetA, when applied on the validation EmotiW data set. We will also show that using (\ref{eq:9}) for NetA training leads the network to better perform on the training set as well.

\section{Experimental Study}

\subsection{The Deep Neural Networks used in the experiments}

The Deep Neural Network architecture proposed in this study, is an end-to-end model including both CNN and RNN components. CNNs are used to extract high level features from the input images, while the RNN exploits the sequential nature of the input data to provide the ﬁnal predictions. 

The CNN subsystem matches the VGG-Face architecture \cite{parkhi2015deep}, a model trained for face verification. Both Bidirectional LSTM and GRU cells were considered as units of the RNN subsystem. The output of the first fully connected layer of the CNN is fed as an input to the RNN, which in turn outputs the ﬁnal prediction of the system.

Table \ref{vggrnn} shows the configuration of the architecture. It is composed of 9 blocks. For each convolutional layer the parameters are denoted as (channels, kernel, stride) and for the max pooling layer as (kernel, stride). It also shows the respective number of units of each fully connected layer.

\begin{table}[h]
\centering
\caption{The CNN-RNN architecture}
\label{vggrnn}\begin{tabular}{|c|c|c|}
\hline
block 1 & $2  \times$ conv layer & (64, $3 \times 3$, $1 \times 1$) \\
&$1 \times$ max pooling & ($2 \times 2$, $2 \times 2$) \\
\hline
block 2 & $2  \times$ conv layer & (128, $3 \times 3$, $1 \times 1$) \\
&$1 \times$ max pooling & ($2 \times 2$, $2 \times 2$) \\
\hline
block 3 & $3 \times$ conv layer & (256, $3 \times 3$, $1 \times 1$) \\
&$1 \times$ max pooling & ($2 \times 2$, $2 \times 2$) \\
\hline
block 4 & $3 \times$ conv layer & (512, $3 \times 3$, $1 \times 1$) \\
&$1 \times$ max pooling & ($2 \times 2$, $2 \times 2$) \\
\hline
block 5 & $3 \times$ conv layer & (512, $3 \times 3$, $1 \times 1$) \\
&$1 \times$ max pooling & ($2 \times 2$, $2 \times 2$) \\
\hline
block 6 &fully connected 1 & 4096\\
&dropout layer&\\
\hline
block 7 & RNN layer 1 & 128\\
&dropout layer&\\
\hline
block 8 & RNN layer 2 & 128\\
\hline
\end{tabular}
\end{table}

\begin{figure}[!ht]
\begin{tabularx}{\linewidth}{>{\hsize=1.2\hsize}X
                             >{\hsize=1.2\hsize}X}
        \includegraphics[width=\hsize,valign=t]{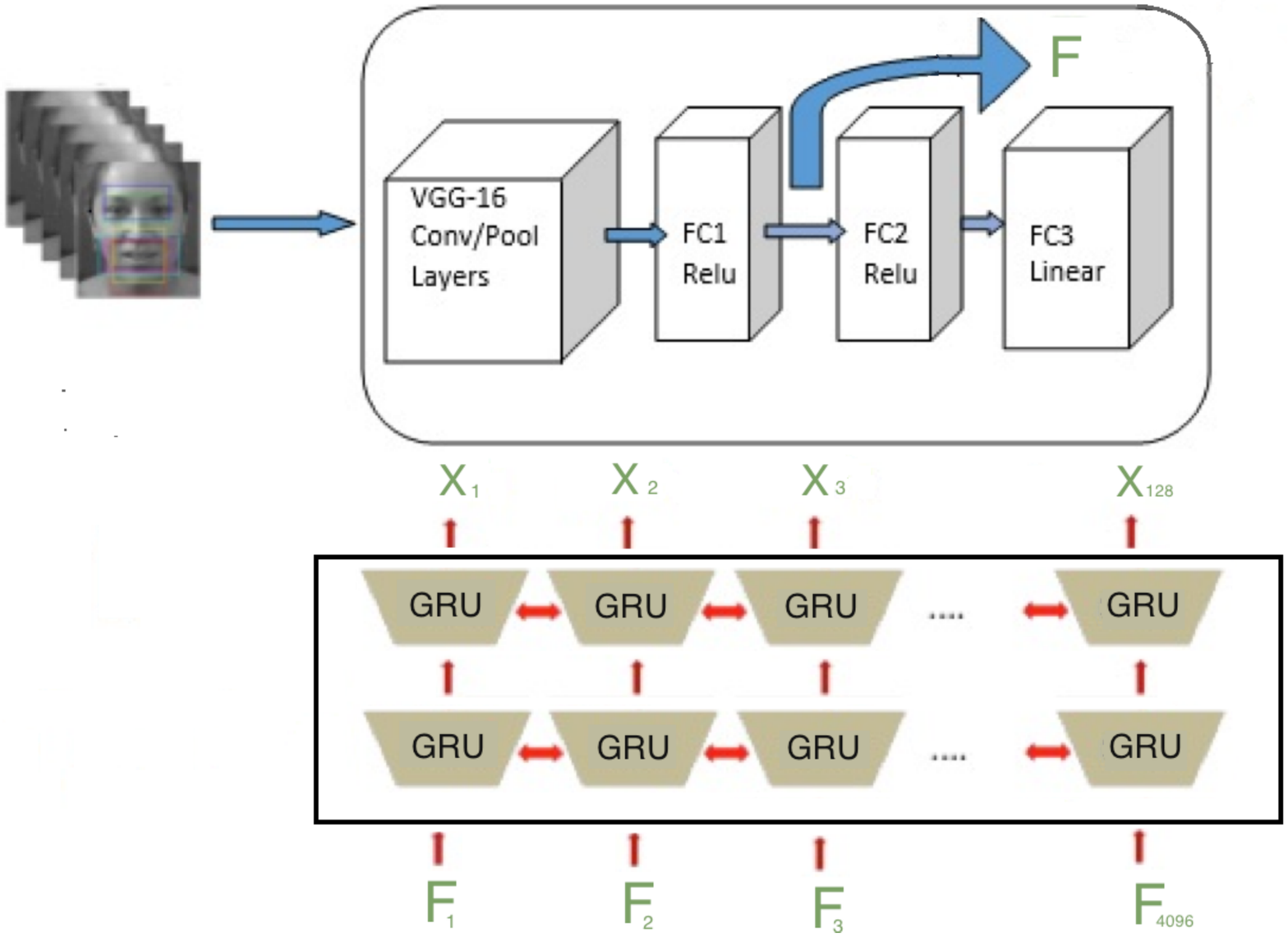}
\caption{ The features F extracted from CNN FC1 layer are passed to the RNN network; the latter provides the final decision}
\label{cnnrnn}
\end{tabularx}
\end{figure}

Fig. \ref{cnnrnn} shows the architecture of the proposed system. The weights from the convolutional and pooling layers of the VGG-Face are initialized from a pre-trained implementation. During the training phase, these parts remain fixed, while only the fully connected layer at the end is actually trained. 

The next layer is a fully connected one, using the Rectified Linear Unit as its activation function. The CNN, using a linear-activated fully connected layer (FC3), can also provide classification to the 7 basic emotion categories. 

The CNN  outputs a vector of 4096 high-level features which are extracted from each video frame (denoted by \textit{F} in Fig. 4). The RNN processes the $F_1$, $F_2$, ... , $F_{4096}$ , corresponding to the number of CNN outputs used and delivers the final representations $X_1$, $X_2$, ..., $X_{128}$, defined in (\ref{eq:4}).

The hyper-parameter values which were used to train the CNN-RNN networks, were selected after extensive experimentation: a sequence length size of 80 consecutive data frames was used so as to provide the RNN part of the network with the ability to detect meaningful time correlations in the data; a constant learning rate of 0.001; 4096 hidden units in the fully connected layer of the CNN part; dropout after this CNN layer with a value of 0.5; 128 hidden units of the GRU type in two RNN layers. The weights were initialized from a Truncated Normal distribution with a zero mean and a variance equal to 0.1, while the biases were initialized to 1. Training was performed on a single GeForce GTX TITAN X GPU.

\subsection{The Training and Validation Data}

As already mentioned, we used the training data of the EmotiW 2017 dataset to train the “NetA” CNN-RNN architecture and the validation set of the same Challenge to train the “NetB” one. 

These two datasets possess quite different characteristics, because they are generated from different video clips in-the-wild. As shown in all three methods that produced the best emotion recognition results in the EmotiW 2017 Grand Challenge \cite{hu2017learning} \cite{knyazev2017convolutional} \cite{vielzeuf2017temporal}, the performances obtained by deep neural networks, which learned the EmotiW training data, on the validation and test data were quite similar. So, our target is to provide an as good as possible performance on the validation set, while achieving the best possible performance on the training set as well.

In our first experiment we trained a deep CNN-RNN network on the validation data. The best trained (NetB) CNN-RNN achieved a training accuracy of 99\%. However, its generalization on the EmotiW 2017 training data was only 33.4\%. Nevertheless, we adopted this configuration as NetB, because obtaining an as high as possible accuracy on the validation data has been crucial for the Challenge goals, as was above mentioned. From this network we extracted the U representations defined in (\ref{eq:1}), for all data in the validation set. About 19,000 frames were 
provided by the Challenge, extracted from the AFEW videos, with the facial areas cropped. The images were resized to the resolution of $96 \times 96 \times 3$. Their values were normalized to the range $[-1, 1]$. We provided all these as inputs to the network. 

Then, we applied \textit{k}-means clustering on these representations, with \textit{k} = 7, defining seven well separated clusters corresponding to the seven basic emotion categories. We derived the cluster centroids defined in (\ref{eq:4}), with \textit{n} = 128  and used these in our following experiments.
Next, we focused our attention on training the NetA network, using the EmotiW train dataset; about 45,000 video frames, extracted, cropped and pre-processed as in the former case, were used for training. 

First, we trained a similar CNN-RNN architecture with the training data, minimizing the error criterion in (\ref{eq:5}) and selected the architecture which provided the best accuracy on the validation data set. This architecture provided a best accuracy of 38,4\% on the validation data, with its performance on the training data being only 60,6\%; illustrating the significant differences between the training and validation data sets.

\subsection{Application of the Proposed Training Procedure}

In the following, we implemented the proposed approach described in Section \ref{sec3}.  
We trained a CNN-RNN architecture minimizing the new error criterion defined in (\ref{eq:9}), using the seven cluster centroids obtained in our former experiment as desired outputs in (\ref{eq:9}) and testing different values of the parameter $\lambda$. Using $\lambda=1$ derives the network described in the former paragraph; a value of  $\lambda=0$ provides a network trying to replicate the 7 cluster centroids at the outputs of its last RNN hidden layer; a value of  $\lambda=0.5$ pays the same attention to both criteria in (\ref{eq:9}).

Table \ref{lambda} summarizes the obtained accuracy on the validation, as well as training data sets, for different values of  $\lambda$. It can be easily shown that the best results have been obtained when $\lambda=0$. The proposed criterion really helped to achieve better results than the original mean squarred error (MSE) criterion. This verifies that the role of validation data is of high importance in this categorical emotion recognition in-the-wild problem.

\begin{table}[h]
\caption{Accuracies of the proposed architecture for the EmotiW training and validation set for different values of $\lambda$}
\label{lambda}
\centering
\begin{tabular}{|c||c|c|}
\hline
\multicolumn{1}{|c||}{Value of $\lambda$} & \multicolumn{2}{|c|}{Accuracy on Set}  \\
\hline
  & Validation & Training \\
\hline
0 & \textbf{0.446} & \textbf{0.69}  \\
\hline
0.25 & 0.426 & 0.67  \\
\hline
0.5 & 0.422 & 0.622  \\
\hline
0.75 & 0.398 & 0.611  \\
\hline
1 & 0.384 & 0.606  \\
\hline
\end{tabular}
\end{table}

Tables \ref{afew} and \ref{afew2} compare the obtained accuracy between $\lambda=0$ and $\lambda=1$ of each category of the validation and training sets, respectively. It can be easily shown that the best results have been obtained when $\lambda=0$. Table \ref{conf_matrix} shows the confusion matrix for the validation set when $\lambda=0$.

\begin{table*}[t]
\caption{Accuracies per category for the EmotiW validation set when  $\lambda=0$ and $\lambda=1$}
\label{afew}
\centering
\begin{tabular}{|c||c|c|c|c|c|c|c|c|}
\hline
\multicolumn{1}{|c||}{Value of $\lambda$} & \multicolumn{8}{|c|}{Accuracy on Validation Set}  \\
\hline
& Neutral & Anger & Disgust & Fear & Happy & Sad & Surprise & Total \\
\hline
0 & \textbf{0.492} & \textbf{0.578} & \textbf{0.075} & \textbf{0.217}  & \textbf{0.667} & \textbf{0.623}  & \textbf{0.217}  & \textbf{0.446} \\
\hline
1  & 0.466 & 0.56 & 0 & 0.046  & 0.635 & 0.431  & 0.111  & 0.384 \\ 
\hline
\end{tabular}
\end{table*}

\begin{table*}[t]
\caption{Accuracies per category for the EmotiW training set when  $\lambda=0$ and $\lambda=1$}
\label{afew2}
\centering
\begin{tabular}{|c||c|c|c|c|c|c|c|c|}
\hline
\multicolumn{1}{|c||}{Value of $\lambda$} & \multicolumn{8}{|c|}{Accuracy on Training Set}  \\
\hline
& Neutral & Anger & Disgust & Fear & Happy & Sad & Surprise & Total \\
\hline
0 & \textbf{0.842} & \textbf{0.854} & \textbf{0.08} & \textbf{0.273}  & \textbf{0.923} & \textbf{0.928}  & \textbf{0.444}  & \textbf{0.69} \\
\hline
1  & 0.833 & 0.802 & 0.007 & 0.014  & 0.901 & 0.707  & 0.164  & 0.606 \\ 
\hline
\end{tabular}
\end{table*}

\begin{table*}[h]
\centering
\caption{Confusion Matrix for the EmotiW validation set when  $\lambda=0$}
\label{conf_matrix}
\begin{tabular}{|c|c|c|c|c|c|c|c|}
\hline
 & Neutral & Anger & Disgust & Fear & Happy & Sad & Surprise  \\
\hline
Neutral & \textbf{0.492}  & 0.111  & 0.032  & 0.047  & 0.175  & 0.111 &  0.032 \\
\hline
Anger & 0.094 & \textbf{0.578}  & 0.108  & 0.063  & 0.016  & 0.078 & 0.063  \\
\hline
Disgust & 0.125 & 0.175  &  \textbf{0.075}  & 0.15  & 0.05  & 0.175 & 0.25  \\
\hline
Fear & 0.109 & 0.109  & 0.152  & \textbf{0.217}  & 0.131  & 0.109 & 0.173  \\
\hline
Happy & 0.175  & 0.016  & 0.016  & 0.016  & \textbf{0.667}  & 0  & 0.11  \\
\hline
Sad & 0.066 & 0.0  & 0.115  & 0.082  & 0.033  & \textbf{0.623}  & 0.081  \\
\hline
Surprise & 0.109  & 0.087  & 0  & 0.261   & 0.174   & 0.152  & \textbf{0.217}   \\
\hline
\end{tabular}
\end{table*}

\begin{table*}[h]
\caption{Total accuracies of the best architectures of the 3 winning methods of the EmotiW 2017 Grand Challenge reported on the validation set vs our own model}
\label{emotiw}
\centering
\begin{tabular}{|c||c|c|c|c|c|}
\hline
\multicolumn{1}{|c||}{Group} & \multicolumn{1}{|c|}{Architecture} & \multicolumn{4}{|c|}{Total Accuracy} \\
\hline
& & Original &  \begin{tabular}{@{}c@{}} SSE Learning \\ Strategy \end{tabular}  & \begin{tabular}{@{}c@{}}After\\ Fine-Tuning\end{tabular}  & \begin{tabular}{@{}c@{}}Data\\ augmentation\end{tabular} \\
\hline
\cite{hu2017learning} &  \begin{tabular}{@{}c@{}} DenseNet-121 \\ HoloNet \\ ResNet-50 \end{tabular}   &   \begin{tabular}{@{}c@{}}  0.414 \\ 0.41 \\ 0.418 \end{tabular} &  \begin{tabular}{@{}c@{}} 0.457 \\ \textbf{0.465} \\ 0.426  \end{tabular} & - & -\\
\hline
\cite{knyazev2017convolutional} &  \begin{tabular}{@{}c@{}} VGG-Face \\ FR-Net-A \\ FR-Net-B \\ FR-Net-C \\ LSTM + FR-NET-B \end{tabular}    &  \begin{tabular}{@{}c@{}} 0.379 \\ 0.337 \\ 0.334 \\ 0.376 \\ - \end{tabular} & - & \begin{tabular}{@{}c@{}}  0.483 \\ 0.446 \\ \textbf{0.488} \\  0.452 \\ 0.465 \end{tabular} &\begin{tabular}{@{}c@{}} - \\ - \\ - \\ - \\ \textbf{0.504} \end{tabular} \\
\hline
\cite{vielzeuf2017temporal}& \begin{tabular}{@{}c@{}} Weighted C3D (no overlap) \\ LSTM C3D (no overlap) \\ VGG-Face \\ VGG-LSTM 1 layer \end{tabular}    &  - & - & - & \begin{tabular}{@{}c@{}} 0.421 \\ 0.432 \\ 0.414 \\ 0.486 \end{tabular}\\
\hline
Our & CNN-RNN with $\lambda=0$   & \textbf{0.446} & - & - & - \\
\hline
\end{tabular}
\end{table*}

It should be mentioned that the performance of the network trained with $\lambda=0$ is much higher than the baseline 38,81\% reported in \cite{dhall2017individual} and better than all other original architectures in the three winning methods in the Audio-video emotion recognition EmotiW 2017 Grand Challenge \cite{hu2017learning} \cite{knyazev2017convolutional} \cite{vielzeuf2017temporal}, as shown in Table \ref{emotiw}.

Here it can be noted that our network was trained to classify only video frames (and not audio) and then video classification based on frame aggregation was performed. Moreover, no data-augmentation, post-processing of the results or ensemble methodology have been conducted, as was done in the above three winning methods. Also, for training the network we used only the cropped faces provided by the challenge. Our results can be even better, if we use the above methodologies, as well as our own detection and/or normalization procedure, but this is not the main scope of this paper.

To illustrate the good performance of the proposed training approach, we computed t-Distributed Stochastic Neighbor Embedding (t-SNE) \cite{hinton2003stochastic}  on: 
\begin{itemize}
\item the seven Z centroids – defined in (\ref{eq:2}), each composed of 128 neuron output values, provided as desired outputs for training NetA; these are marked with dots,
\item the seven centroids, computed by similarly clustering all internal representations X - defined in (\ref{eq:4}) and composed of 128 elements as well - extracted from NetA, during its training through minimization of the error in (\ref{eq:9}), with $\lambda=0$ ; these are marked with stars,
\item the seven respective centroids computed by clustering the internal representations of the network trained through minimization of the error in (\ref{eq:5}) or equivalently in (\ref{eq:9}) with $\lambda=1$; these are marked with crosses.  
\end{itemize}

\begin{figure}[!ht]
\centering
\adjincludegraphics[height=7cm,trim={.50\width} {.30\width} {.50\width} {.30\width}]{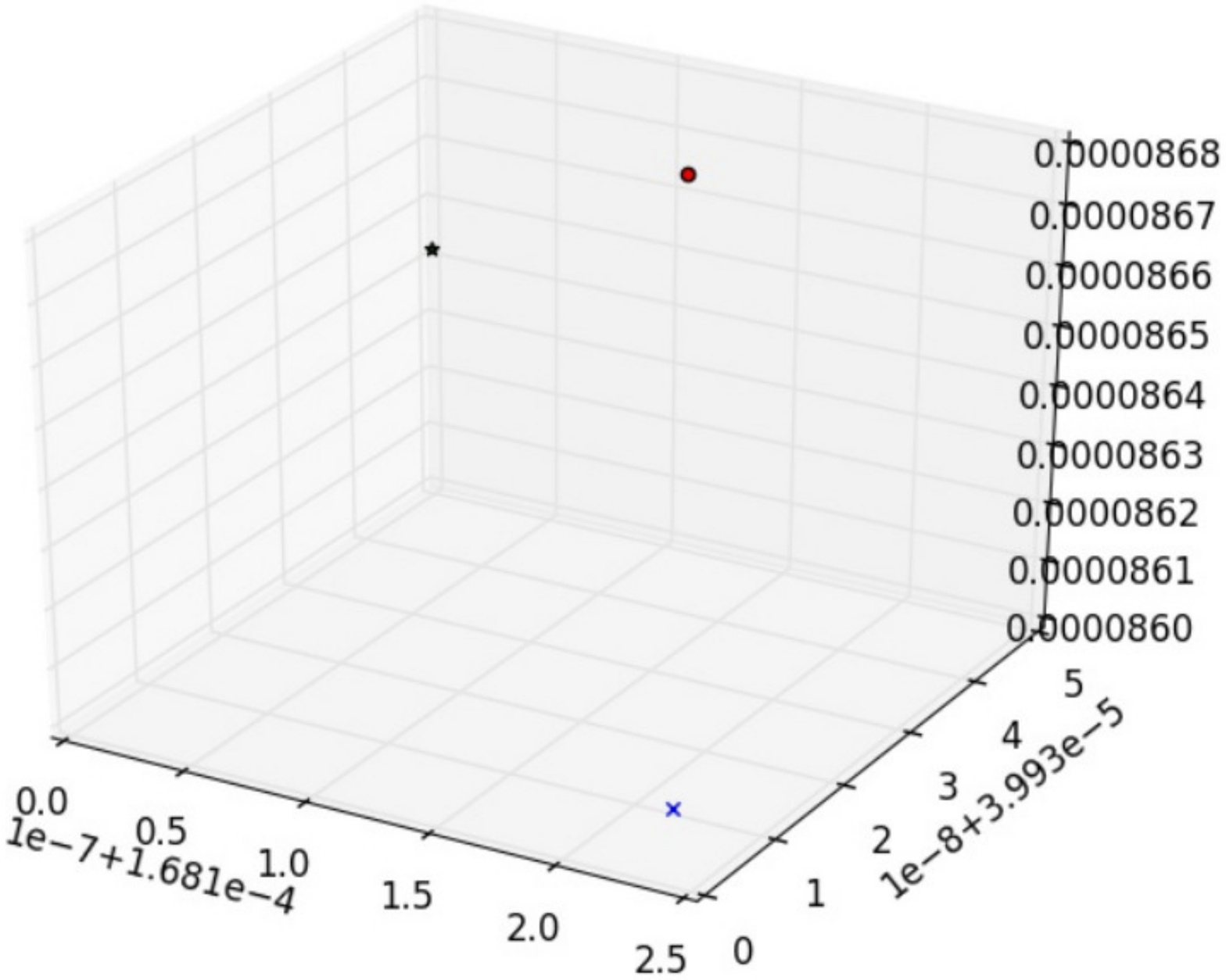} 
\caption{ Improving Neutral cluster centroid proximity through (\ref{eq:9}) }
\label{neutral}
\end{figure}

\begin{figure}[!ht]
\centering
\adjincludegraphics[height=7cm,trim={.50\width} {.30\width} {.50\width} {.30\width}]{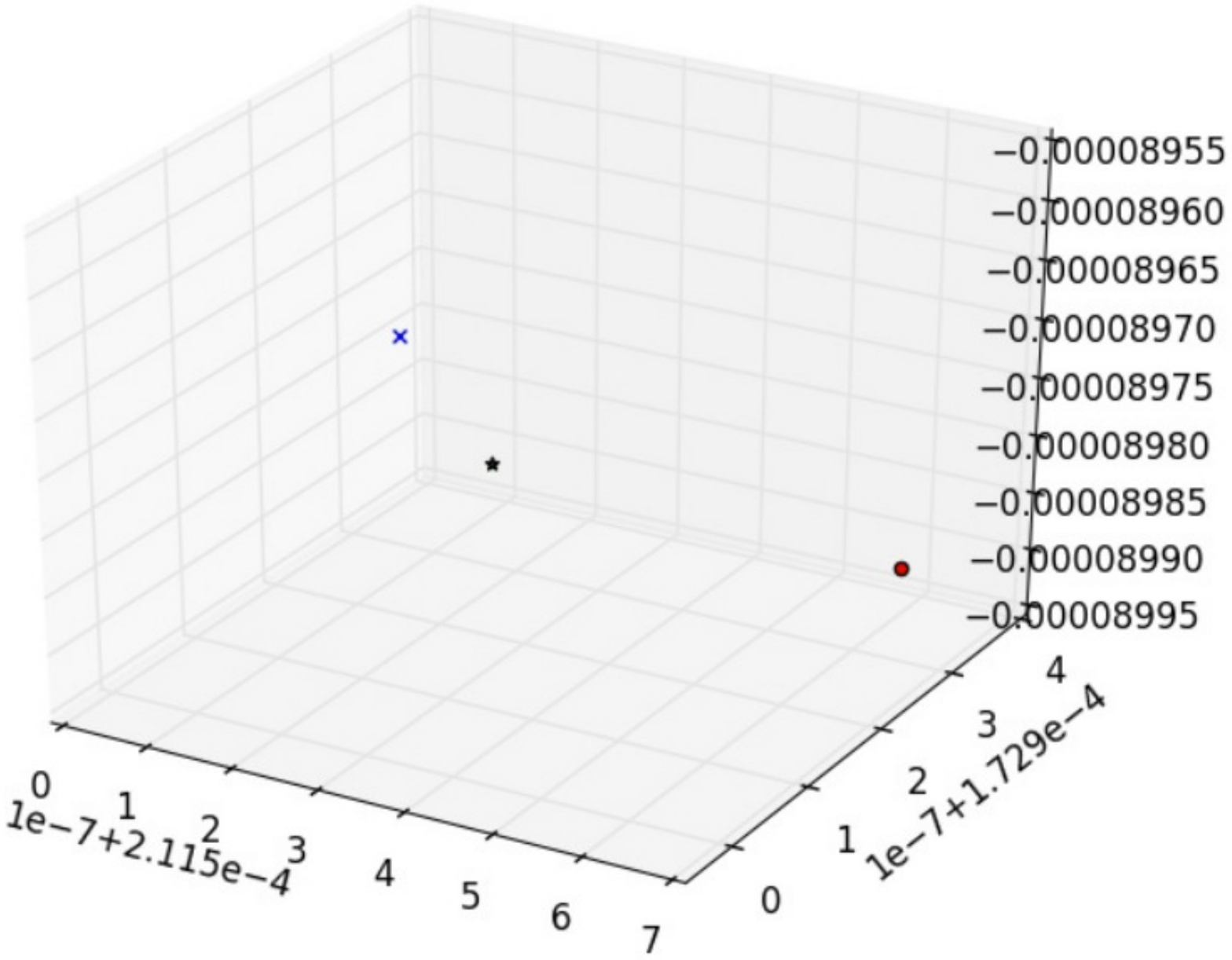} 
\caption{ Improving Anger cluster centroid proximity through (\ref{eq:9}) }
\label{anger}
\end{figure}

\begin{figure}[!ht]
\centering
\adjincludegraphics[height=7cm,trim={.50\width} {.30\width} {.50\width} {.30\width}]{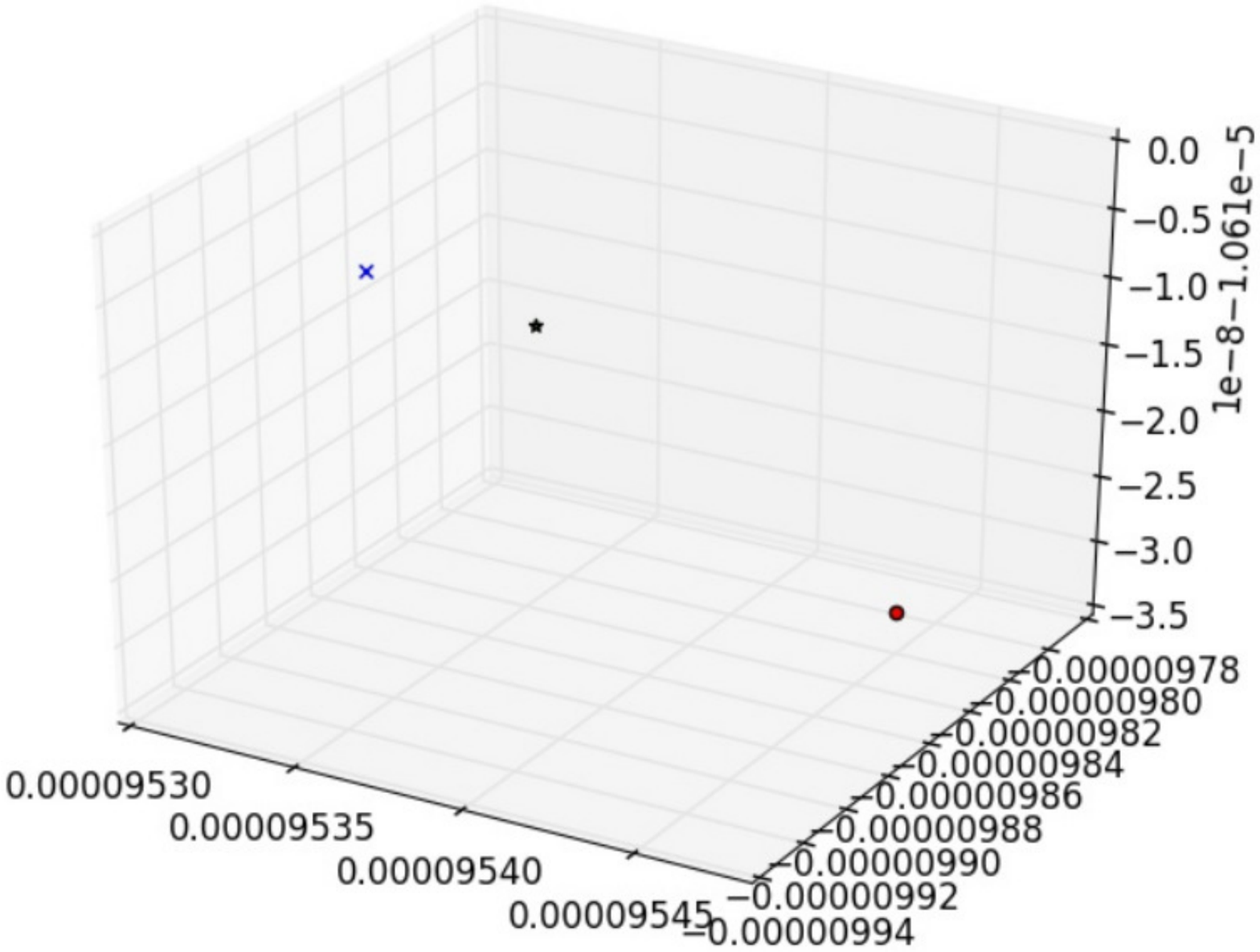} 
\caption{ Improving Disgust cluster centroid proximity through (\ref{eq:9}) }
\label{disgust}
\end{figure}

\begin{figure}[!ht]
\centering
\adjincludegraphics[height=7cm,trim={.50\width} {.30\width} {.50\width} {.30\width}]{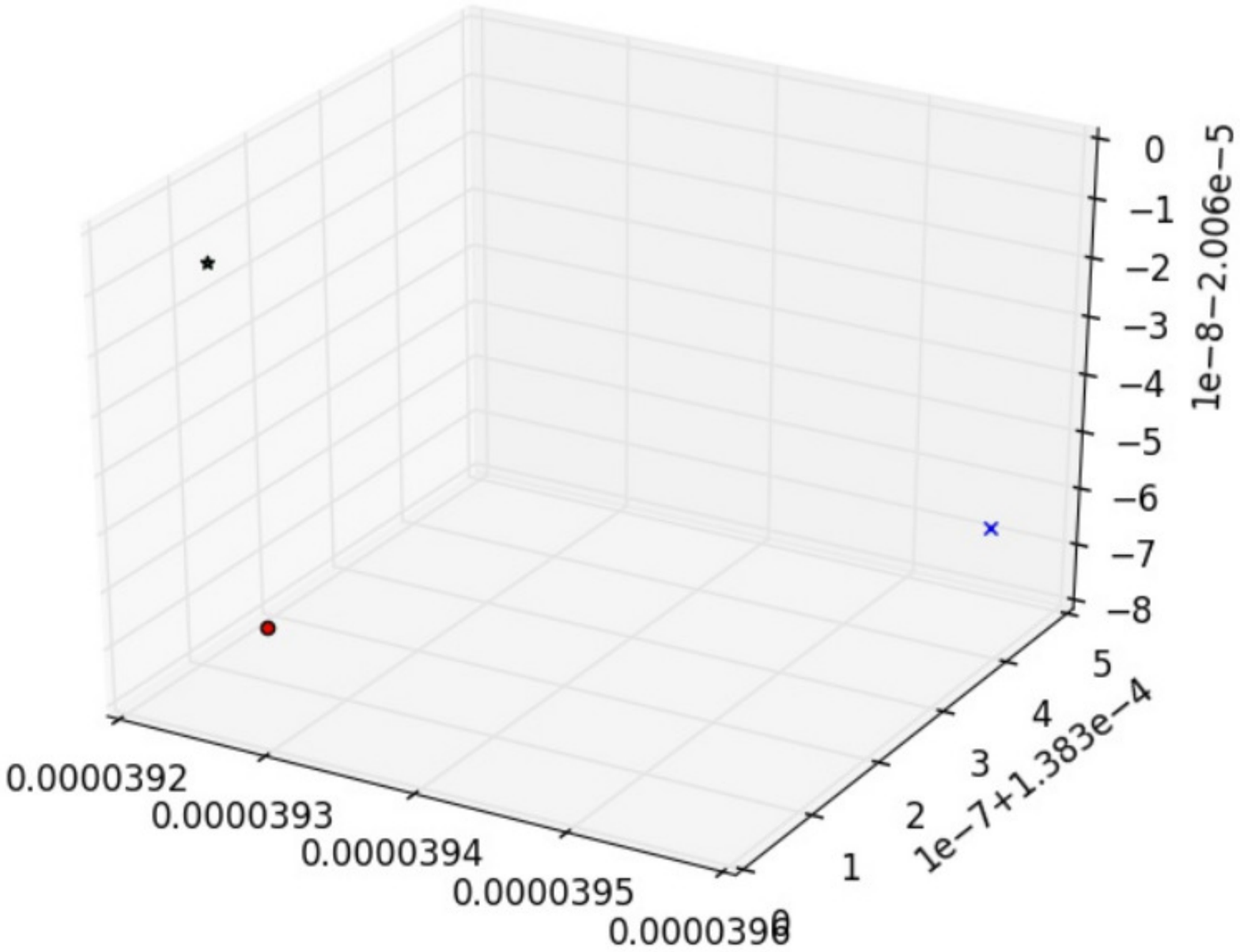} 
\caption{ Improving Sad cluster centroid proximity through (\ref{eq:9}) }
\label{sad}
\end{figure}

\begin{figure}[!ht]
\centering
\adjincludegraphics[height=7cm,trim={.50\width} {.30\width} {.50\width} {.30\width}]{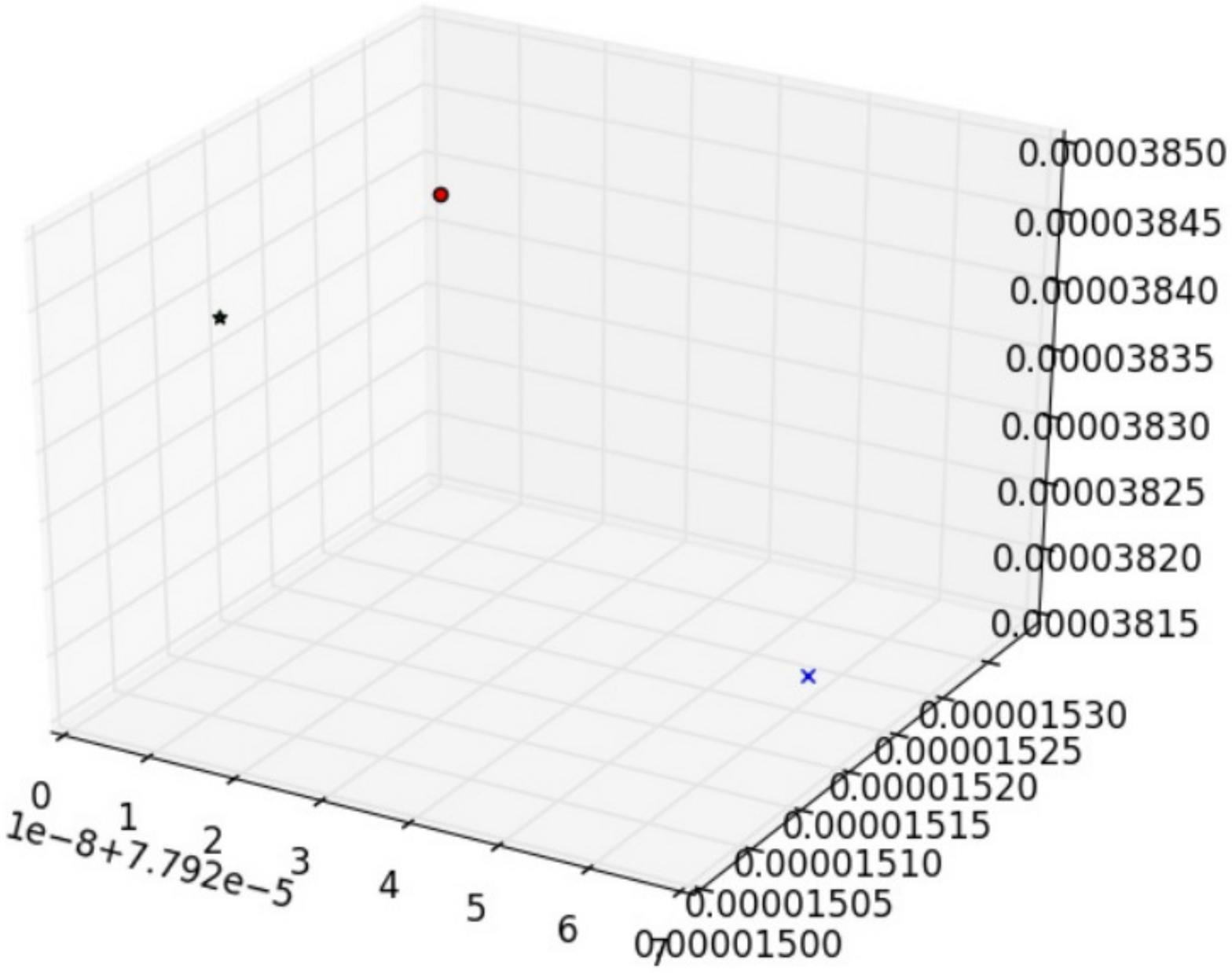} 
\caption{ Improving Happy cluster centroid proximity through (\ref{eq:9}) }
\label{happy}
\end{figure}

Figs. \ref{neutral}-\ref{happy} show the respective cluster centroids, visualized in a three-dimensional space, for some of the emotion categories. 
Table \ref{lambda2} shows: i) the mean distance of each of the centroids marked with stars from the respective Z centroids marked with dots ($\lambda=0$) and ii) the mean distance of each of the centroids marked with crosses from the respective Z centroids marked with dots ($\lambda=1$). 

It can be easily verified that in all cases NetA trained with the proposed procedure manages to bring the cluster centroids, corresponding to the EmotiW training data, closer to the respective cluster centroids of the EmotiW validation dataset. This leads to achieving higher accuracies in both datasets.

\begin{table}
\caption{Distances between training and validation cluster centers}
\label{lambda2}
\centering
\begin{tabular}{|c||c|c|}
\hline
\multicolumn{1}{|c||}{Category } & \multicolumn{2}{|c|}{Mean Distance from respective validation cluster center}  \\
\hline
  & $\lambda=0$ & $\lambda=1$ \\
\hline
Neutral & 0.015 & 0.067  \\
\hline
Anger & 0.015 &0.061  \\
\hline
Disgust & 0.052 & 0.058  \\
\hline
Fear & 0.055& 0.065  \\
\hline
Happy & 0.012 &0.077  \\
\hline
Sad & 0.025& 0.065  \\
\hline
Surprise & 0.055 & 0.067  \\
\hline
\end{tabular}
\end{table}

\section{Conclusions and Further Work}

In this paper we have defined a new error criterion for training a deep neural network with different datasets in-the-wild, so as to achieve a high network performance in both datasets. We used the categorical emotion recognition paradigm, using datasets collected from movies and tv clips, as a testbed where the proposed approach can provide improved deep neural network performances.

In particular, we extracted internal representations at the final hidden layer of a deep CNN-RNN architecture and created seven clusters corresponding to the primary emotion categories, i.e. anger, disgust, fear, happiness, sadness, surprise and the neutral state. We then used the above cluster centroids as desired outputs for training a new deep CNN-RNN with another dataset for emotion recognition in-the-wild. We defined an appropriate loss function to be used for this domain adaptation procedure.

We were able to illustrate, through an experimental study, that the latter network was able to have an improved performance on its training set by about 14\%, as well as on the data set of the former network, by about 12\%.

Compared to transfer learning or retraining of a DNN with the new dataset, the proposed approach provides the following advantages: i) it reduces catastrophic forgetting, since the desired outputs in DNN training also include the cluster centroids of the former dataset and ii) much less resources are necessary, since our method does not require availability of the former DNN architecture/weights, nor of any large amount of former data.

In our future work, the proposed procedure will be extended, to obtain a block component form, in which the proposed error criterion is iteratively minimized by each of the above networks.
Future work also includes extending the proposed approach to deal with dimensional emotion recognition for estimation of the Valence and Arousal values. To do so, we will discretize the 2D Valence Arousal Space, e.g. in 400 (20 $\times$ 20) classes and then apply the proposed approach between different in-the-wild datasets, such as AffWild, RECOLA \cite{ringeval2013introducing} and OMG-Emotion Behavior Dataset \cite{barros2018omg}.

\section{Acknowledgment}

The work of Stefanos Zafeiriou has been partially funded by the FiDiPro program of Tekes (project number: 1849/31/2015). The work of Dimitris Kollias was funded by a Teaching Fellowship of Imperial College London. We also wish to thank Dr Abhinav Dall for providing us with the datasets of all recent EmotiW Challenges~\cite{dhall2017individual}-~\cite{dhall2015video}.

\bibliographystyle{spmpsci}      
\bibliography{sample}   

\end{document}